\documentclass[12pt,twoside]{article}
\newdimen\paravsp  \paravsp=1.3ex 
\usepackage{amsmath,amssymb}
\def\,{\mskip 3mu} \def\>{\mskip 4mu plus 2mu minus 4mu} \def\;{\mskip 5mu plus 5mu} \def\!{\mskip-3mu}
\def\dispmuskip{\thinmuskip= 3mu plus 0mu minus 2mu \medmuskip=  4mu plus 2mu minus 2mu \thickmuskip=5mu plus 5mu minus 2mu}
\def\textmuskip{\thinmuskip= 0mu                    \medmuskip=  1mu plus 1mu minus 1mu \thickmuskip=2mu plus 3mu minus 1mu}
\textmuskip
\def\beq{\dispmuskip\begin{equation}}    \def\eeq{\end{equation}\textmuskip}
\def\beqn{\dispmuskip\begin{displaymath}}\def\eeqn{\end{displaymath}\textmuskip}
\def\bqa{\dispmuskip\begin{eqnarray}}    \def\eqa{\end{eqnarray}\textmuskip}
\def\bqan{\dispmuskip\begin{eqnarray*}}  \def\eqan{\end{eqnarray*}\textmuskip}

\newtheorem{theorem}{Theorem}
\newtheorem{open}[theorem]{Open Problem}
\newenvironment{keywords}{\centerline{\bf\small
Keywords}\begin{quote}\small}{\par\end{quote}\vskip 1ex}

\def\paradot#1{\vspace{\paravsp plus 0.5\paravsp minus 0.5\paravsp}\noindent{\bf\boldmath{#1.}}} 
\def\paranodot#1{\vspace{\paravsp plus 0.5\paravsp minus 0.5\paravsp}\noindent{\bf\boldmath{#1}}} 
\def\qmbox#1{{\quad\mbox{#1}\quad}} 
\def\req#1{\eqref{#1}}          
\def\eps{\varepsilon}           
\def\epstr{\epsilon}            
\def\nq{\hspace{-1em}}          
\def\qed{\hspace*{\fill}\rule{1.4ex}{1.4ex}$\quad$\\} 
\def\frs#1#2{{^{#1}\!/\!_{#2}}} 
\def\fr#1#2{{\textstyle{#1\over#2}}} 
\def\SetN{\mathbb{N}}           
\def\SetB{\mathbb{B}}           
\def\SetQ{\mathbb{Q}}           
\def\e{{\rm e}}                 
\def\v{\boldsymbol}             
\def\E{{\mathbb E}}             
\def\g{\gamma}
\def\d{\delta}
\def\A{{\cal A}}                
\def\X{{\cal X}}                
\def\M{{\cal M}}                
\def\N{{\cal N}}                
\def\P{{\cal P}}                
\def\t{\tilde}                  
\def\Srat{\text{rat}}
\def\Sno{\text{n1}}
\def\Slim{\text{lim}}
\def\Smix{\text{mix}}
\def\DTIME{\text{TIME}}

\begin{document}

\title{\vspace{-4ex}
\vskip 2mm\bf\Large\hrule height5pt \vskip 4mm
Offline to Online Conversion
\vskip 4mm \hrule height2pt}
\author{{\bf Marcus Hutter}\\[3mm]
\normalsize Research School of Computer Science \\[-0.5ex]
\normalsize Australian National University \\[-0.5ex]
\normalsize Canberra, ACT, 0200, Australia \\
\normalsize \texttt{http://www.hutter1.net/}
}
\date{11 July 2014}
\maketitle

\begin{abstract}
We consider the problem of converting offline estimators into an
online predictor or estimator with small extra regret. Formally
this is the problem of merging a collection of probability
measures over strings of length 1,2,3,... into a single
probability measure over infinite sequences. We describe various
approaches and their pros and cons on various examples. As a
side-result we give an elementary non-heuristic purely
combinatoric derivation of Turing's famous estimator. Our main
technical contribution is to determine the computational
complexity of online estimators with good guarantees in general.
\vspace*{1ex}
\def\contentsname{\centering\normalsize Contents}\setcounter{tocdepth}{1}
{\parskip=-2.7ex\tableofcontents}
\end{abstract}


\newpage
\section{Introduction}\label{sec:Intro}

A standard problem in statistics and machine learning is to
estimate or learn an in general non-i.i.d.\ probability
distribution $q_n:\X^n\to[0,1]$ from a batch of data $x_1,...,x_n$.
$q_n$ might be the Bayesian mixture over a class of distributions
$\M$, or the (penalized) maximum likelihood (ML/MAP/MDL/MML)
distribution from $\M$, or a combinatorial probability, or an
exponentiated code length, or else. This is the batch or {\em
offline} setting. An important problem is to predict $x_{n+1}$ from
$x_1,...,x_n$ sequentially for $n=0,1,2...$, called {\em online}
learning if the predictor improves with $n$. A stochastic
prediction $\t q(x_{n+1}|x_{1:n})$ can be useful in itself (e.g.\
weather forecasts), or be the basis for some decision, or be used
for data compression via arithmetic coding, or otherwise. We use
the prediction picture, but could have equally well phrased
everything in terms of log-likelihoods, or perplexity, or
code-lengths, or log-loss.

The naive predictor is $\t
q^\Srat(x_{n+1}|x_1...x_n):=q_{n+1}(x_1...x_{n+1})/q_n(x_1...x_n)$
is not properly normalized to 1 if $q_n$ and $q_{n+1}$ are not
compatible. We could fix the problem by normalization $\t
q^\Sno(x_{n+1}|x_1...x_n):=\t
q^\Srat(x_{n+1}|x_1...x_n)/\sum_{x_{n+1}}\t
q^\Srat(x_{n+1}|x_1...x_n)$, but this may result in a very poor
predictor. We discuss two further schemes, $\t q^\Slim$ and $\t
q^\Smix$, the latter having good performance guarantees (small
regret), but a direct computation of either is prohibitive.

A major open problem is to find a computationally tractable online
predictor $\t q$ with provably good performance given offline
probabilities ($q_n$). A positive answer would benefit many applications.

\paradot{Applications}
(i) Being able to use an offline estimator to make stochastic
predictions (e.g.\ weather forecasts) is of course useful. The
predictive probability needs to sum to 1 which $\t q^\Sno$
guarantees, but the regret should also be small, which only $\t
q^\Smix$ guarantees.

(ii) Given a parameterized class of (already) online estimators
$\{\t q^\theta\}$, estimating the parameter $\theta$ from data $x_1...x_n$
(e.g. maximum likelihood) for $n=1,2,3,...$ leads to a sequence of
parameters $(\hat\theta_n)$ and a sequence of estimators
$(q_n):=(\t q^{\hat\theta_n})$ that is usually {\em not} online. They
need to be reconverted to become online to be useful for prediction or
compression, etc.

(iii) Arithmetic coding requires an online estimator, but often is based
on a class of distributions as described in (ii). The default
`trick' to get a fast and online estimator is to use
$\t q^{\hat\theta_n}(x_{n+1}|x_{1:n})$ which is properly
normalized and often very good.

(iv) Online conversions are needed even for some offline purposes.
For instance, computing the cumulative distribution function
$\sum_{y_{1:n}\leq x_{1:n}} q_n(y_{1:n})$ can be hard in general,
but can be computed in time $O(n)$ if $(q_n)$ is (converted to) online.

\paradot{Contributions \& contents}
The main purpose of this paper is to introduce and discuss the
problem of converting offline estimators $(q_n)$ to an online
predictor $\t q$ (Section~\ref{sec:Form}).

We compare and discuss the pros and cons of the four conversion proposals (Section~\ref{sec:Sol}).
We also define the worst-case extra regret of online $\t q$ over offline $(q_n)$,
measuring the conversion quality.

We illustrate their behavior for various classical estimators (Bayes, MDL, Laplace, Good-Turing,
Ristad) (Section~\ref{sec:Ex}). Naive normalization of the triple uniform
estimator interestingly leads to the Good-Turing estimator, but induces huge extra regret,
while naive normalization of Ristad's quadruple uniform estimator induces negligible extra regret.

Given that $\t q^\Sno$ can fail for interesting offline estimators,
natural questions to ask are: whether the excellent predictor $\t q^\Smix$
can be computed or approximated (yes), by an efficient algorithm (no),
whether for every $(q_n)$ there exists any fast $\t q$ nearly as good as
$\t q^\Smix$ (no), or whether there exist $(q_n)$ for which no fast
$\t q$ can even slightly beat the trivial uniform predictor (yes)
(Section~\ref{sec:CCqt}).

The proofs for these computational complexity results are deferred to
the next section (Section~\ref{sec:CCqtproofs}).

These results do not preclude a satisfactory positive solution in
practice, in particular given the contrived nature of the
constructed $(q_n)$, but as any negative complexity result they show
that a solution requires extra assumptions or to moderate our
demands. This leads to some precise open problems to this effect
(Section~\ref{sec:Open}).

Proofs for the regret bounds can be found in Appendix~\ref{app:RnGT} and
a list of notation in Appendix~\ref{app:Notation}.

As a side-result we give the arguably most convincing (simplest and
least heuristic) derivation of the famous Good-Turing estimator.
Other attempts at deriving the estimator Alan Turing suggested in
1941 to I.J.~Good are less convincing (to us) \cite{Good:53}. They
appear more heuristic or convoluted, or are incomplete, often
assuming something close to what one wants to get out
\cite{Nadas:85}. Our purely combinatorial derivation also feels
right for 1941 and Alan Turing.

\section{Problem Formulation}\label{sec:Form}

We now formally state the problem of offline to online conversion
in three equivalent ways and the quality of a conversion.
Let $x_t\in\X$ for $t\in\{1,...,n\}$ and $x_{t:n}:=x_t...x_n\in\X^{n-t+1}$,
$x_{<n}:=x_1...x_{n-1}\in\X^{n-1}$, and $x_{1:0}=x_{<1}=\epstr$ be the empty string.
$\ln$ denotes the natural logarithm and $\log$ the binary logarithm.
$\t q_{|\X^n}$ constrains the domain $\X^*$ of $\t q$ to $\X^n$.

\paradot{Formulation 1 (measures)}
Given probability measures $Q_n$ on $\X^n$ for $n=1,2,3,...$,
find a probability measure $\t Q$ on $\X^\infty$ close to all $Q_n$
in the sense of $\t Q(\A\times\X^\infty)\approx Q_n(\A)$ for all measurable
$\A\subseteq\X^n$ and all $n$.

For simplicity of notation, we will restrict to countable $\X$,
and all examples will be for finite $\X=\{1,...,d\}$.
This allows us to reformulate the problem in terms of probability (mass) functions
and predictors. A choice for $\approx$ will be given below.

\paradot{Formulation 2 (probability mass function)}
Given probability mass functions $q_n:\X^n\to[0;1]$, i.e.\
$\sum_{x_{1:n}}q_n(x_{1:n})=1$, find a function $\t q:\X^*\to[0;1]$
which is {\em time-consistent} \req{TC} in the sense
\beq\label{TC}\tag{TC}
  \sum_{x_n}\t q(x_{1:n}) ~=~ \t q(x_{<n})~\forall n,x_{<n} \qmbox{and} \t q(\epstr)=1
\eeq
and is close to $q_n$ i.e.\ $\t q(x_{1:n})\approx q_n(x_{1:n})$ for all $n$ and $x_{1:n}$.

This is equivalent to Formulation 1, via
$q_n(x_{1:n}):=Q_n(\{x_{1:n}\})$, and since $\t q$ is \ref{TC} iff
there exists $\t Q$ with $\t q(x_{1:n})=\t
Q(\{x_{1:n}\}\times\X^\infty)$
\cite[Appendix]{Hutter:14martoscx}. We will use the following
equivalent predictive formulation, discussed in the introduction,
whenever convenient:

\paradot{Formulation 3 (predictors)}
Given $q_n$ as before, find a predictor $\t q:\X\times\X^*\to[0;1]$
which must be {\em normalized} as
\beq\label{Norm}\tag{Norm}
  \smash{\sum_{x_n}\t q(x_n|x_{<n}) ~=~ 1~~\forall n,x_{<n}}
\eeq
such that its joint probability
\beqn
  \t q(x_{1:n}) ~:= \prod_{t=1}^n\t q(x_t|x_{<t})
\eeqn
is close to $q_n$ as before.

$\t q(x_{1:n})$ is the probability that an (infinite) sequence
starts with $x_{1:n}$ and $\t q(x_n|x_{<n})\equiv\t q(x_{1:n})/\t
q(x_{<n})$ is the probability that $x_n$ follows given $x_{<n}$.
Conditions \req{TC} and \req{Norm} are equivalent, and are the formal
requirement(s) for an estimator to be {\em online}.
We also speak of $(q_n)$ being (not) \ref{Norm} or \ref{TC}.

\paradot{Performance/distance measure}
For modelling and coding we want $\t q$ as large as possible, which suggests
the worst-case regret or log-loss regret
\beq\label{eq:Rndef}
  R_n ~\equiv~ R_n(\t q) ~\equiv~ R_n(\t q||q_n)
  ~:=~ \max_{x_{1:n}}\ln{q_n(x_{1:n})\over\t q(x_{1:n})}
\eeq
For our qualitative considerations, other continuous $R_n\geq 0$
with $R_n=0$ iff $\t q_{|\X^n}=q_n$ would also do. The $R_n$
quantification of $\approx$ above has several convenient properties:
Since an online arithmetic code of $x_{1:n}$ w.r.t.\ $\t q$ has
code length $|\log_2\t q(x_{1:n})|$, and an offline Shannon-Fano
or Huffman code for $x_{1:n}$ w.r.t.\ $q_n$ has code length $|\log_2
q_n(x_{1:n})|$, this shows that the online coding of $x_{1:n}$
w.r.t.\ $\t q$ leads to codes at most $R_n\ln 2$ bits longer than
offline codes w.r.t.\ $q_n$. Naturally we are interested in $\t q$
with small $R_n$, and indeed we will see that this is always achievable.
Also, if $q_n$ is an offline approximation of the true sampling
distribution $\mu$, then $R_n$ upper bounds the {\em extra regret} of a
corresponding online approximation $\t q$:
\beq\label{eq:ExR}
  R_n^{\text{online}}-R_n^{\text{offline}}
  ~\equiv~ R_n(\t q||\mu)-R_n(q_n||\mu)
  ~\leq~ R_n(\t q||q_n)
  ~\equiv~ R_n
\eeq

\paradot{Extending $q_s$ from $\X^s$ to $\X^\infty$}
Some (natural) offline $(q_n)_{n\in\SetN}$ considered later are
automatically online in the sense that $\t q$ defined by $\t
q(x_{1:n}):=q_n(x_{1:n})$ $\forall n,x_{1:n}$ is \ref{TC} and hence
$R_n=0$ for all $n$.
Note that it is {\em always} possible to choose $\t
q$ such that $R_n=0$ for {\em some} $n$: For some fixed $s\in\SetN_0$ define
\beq\label{eq:qbar}
  \bar q_s(x_{1:n}) ~:=~ \left\{
  \begin{array}{ccl}
    q_s(x_{1:s}) & \text{if} & n=s, \\
    \sum_{x_{n+1:s}}q_s(x_{1:s}) & \text{if} & n<s, \\
    q_s(x_{1:s})Q(x_{s+1:n}|x_{1:s}) & \text{if} & n>s
  \end{array}
  \right.
\eeq
where $Q$ can be an arbitrary measure on $\X^\infty$, e.g.\ uniform $Q(x_{s+1:n}|x_{1:s})=|\X|^{n-s}$.
It is easy to see that $\t q:=\bar q_s$ is \ref{TC} with $R_s(\t q)=R_s(\bar q_s)=R_s(q_s)=0$,
but in general $R_n(\bar q_s)>0$ for $n\neq s$.
Therefore naive minimization of $R_n$ w.r.t.\ $\t q$ does not work.
Minimizing $\lim_{n\to\infty} R_n$ can also fail for a number of reasons:
the limit may not exist or is infinite,
or minimizing it leads to poor finite-$n$ performance
or is not analytically possible or computationally intractable.

\section{Conversion Methods}\label{sec:Sol}

We now consider four methods of converting offline estimators to
online predictors and discuss their pros and cons. They illustrate
the difficulties and serve as a starting point to a more
satisfactory solution.

\paradot{Naive ratio}
The simplest way to define a predictor $\t q$ from $q_n$ is via {\em rat}io
\beq\label{eq:qrat}
  \t q^\Srat(x_t|x_{<t}) ~:=~ {q_t(x_{1:t})\over q_{t-1}(x_{<t})}
  \qmbox{or equivalently} \t q^\Srat(x_{1:n}) ~:=~ q_n(x_{1:n})
\eeq
While this ``solution'' is tractable, it obviously only works
when $q_n$ already is \ref{TC}. Otherwise $\t q^\Srat$ violates \req{TC}.
The deviation of
\beq\label{def:norm}
  \N(x_{<t}) ~:=~ \sum_{x_t} \t q^\Srat(x_t|x_{<t}) ~\equiv~ {\sum_{x_t} q_t(x_{1:t})\over q_{t-1}(x_{<t})}
\eeq
from 1 measures the degree of violation. Note that the expectation of
$\N(x_{<t})$ w.r.t. $q_{t-1}$ is 1, so if $\N(x_{<t})$ is smaller than 1 for some $x_{<t}$ it must be larger for others,
hence $\max_{x_{<t}}\N(x_{<t})=1$ iff $\N(x_{<t})=1$ for all $x_{<t}\in\X^{t-1}$.

\paradot{Naive normalization}
Failure of $\t q^\Srat(x_t|x_{<t})$ to satisfy \req{Norm} is easily corrected by normalization \cite{Solomonoff:78}:

\bqa
  \label{eq:qn1c} \t q^\Sno(x_t|x_{<t}) &:=& {q_t(x_{1:t})\over\sum_{x_t} q_t(x_{1:t})}
  ~\equiv~ {\t q^\Srat(x_t|x_{<t})\over\N(x_{<t})} \qmbox{and} \\
  \label{eq:qn1} \t q^\Sno(x_{1:n}) &:=& \prod_{t=1}^n \t q^\Sno(x_t|x_{<t}) ~\equiv {q_n(x_{1:n})\over\prod_{t=1}^n\N(x_{<t})}
\eqa
This guarantees \ref{TC} and for small $\X$ is still tractable, but note
that $\t q^\Sno_{|\X^n}\not\equiv q_n$ unless $q_n$ is already \ref{TC}.
Unfortunately, this way of normalization can result in poor
performance and very large regret $R_n$ for finite $n$ and
asymptotically. Even if performance
is good, computing $R_n$ or finding good upper bounds can be
very hard. Using \req{eq:Rndef} and \req{eq:qn1},
the regret can be represented and upper bounded as follows:
\beq\label{eq:Rnqno}
  R_n(\t q^\Sno) ~=~ \max_{x_{1:n}}\sum_{t=1}^n\ln\N(x_{<t})
  ~\leq~ \sum_{t=1}^n\ln\max_{x_{<t}}\N(x_{<t})
\eeq
If $q_n$ is \ref{TC}, then $\N\equiv 1$, hence $R_n$ as well as the upper bound are 0.

Let us consider here a simple but artificial example how bad things can get,
following up with important practical examples in the next section.
For an i.i.d.\ estimator $q_n(x_{1:n})=q_n(x_1)\cdot...\cdot q_n(x_n)$,
where we slightly overloaded notation, $\t q^\Sno(x_t|x_{<t})=q_t(x_t)$ and
$\t q^\Sno(x_{1:n})=q_1(x_1)\cdot...\cdot q_n(x_n)$, therefore by definition \req{eq:Rndef}
\beqn
  R_n(\t q^\Sno) ~=~ \max_{x_{1:n}}\ln\prod_{t=1}^n {q_n(x_t)\over q_t(x_t)}
  ~=~ \sum_{t=1}^n\ln\max_{x_t} {q_n(x_t)\over q_t(x_t)}
\eeqn
We now consider $\X=\{0,1\}$ with concrete Bernoulli($\frs23$) probability $q_n(x_t=1)=\frs23$ for even $n$
and Bernoulli($\frs13$) probability $q_n(x_t=1)=\frs13$ for odd $n$.
We see that for even $t$,
\beqn
  \t q^\Srat(1_t|1_{<t}) ~=~ {q_t(1_1)\cdot...\cdot q_t(1_{t-1})\cdot q_t(1_t)\over q_{t-1}(1_1)\!\cdot\!...\!\cdot\!q_{t-1}(1_{t-1})~~~~~~~~~~}
  ~=~ 2^{t-1}\cdot{2\over 3}
\eeqn
is very badly unnormalized. Indeed $R_n(\t q^\Sno)$ grows linearly with $n$, i.e.\ becomes very large:
\beqn
  R_n(\t q^\Sno) ~=~ \sum_{t=1}^n\ln
  \left\{ {1 ~\text{ if }~ n-t \text{ is even} \atop
           2 ~\text{ if }~ n-t \text{ is odd } } \right\}
  ~=~ \lfloor{n\over 2}\rfloor \ln 2
\eeqn

\paradot{Limit}
We have seen how to make $R_s=0$ for any fixed $s$ using $\bar q_s$ \req{eq:qbar}.
A somewhat natural idea is to define
\beqn
  \t q^\Slim(x_{1:n}) ~:=~ \lim_{s\to\infty} \bar q_s(x_{1:n}) ~=~
  \lim_{s\to\infty} \sum_{x_{n+1:s}}q_s(x_{1:s})
\eeqn
in the hope to make $\lim_{s\to\infty} R_s=0$. Effectively what $\t
q^\Slim$ does is to use $q_s$ for very large $s$ also for short
strings of length $n$ by marginalization.
Problems are plenty: The limit may not exist, may exist but be incomputable,
$R_n$ may be hard to impossible to compute or upper bound,
and even if the limit exists, $\t q^\Slim$ may perform badly.

For instance, for the above Bernoulli($\frs13|\frs23$) example, the argument of the limit
\beqn
  \t q^\Slim(x_{1:n}) ~=~ \lim_{s\to\infty} \sum_{x_{n+1:s}} q_s(x_1)\cdot...\cdot q_s(x_s)
  ~=~ \lim_{s\to\infty} [q_s(x_1)\cdot...\cdot q_s(x_n)]
\eeqn
oscillates indefinitely (except if $x_1+...+x_n=\frs{n}{2}$).
A template leading to a converging but badly performing $\t q^\Slim$
is $q_n(x_{1:n})=\text{Bad}(x_{<\lfloor n/2\rfloor})\cdot\text{Good}(x_{\lfloor n/2\rfloor:n})$.
While offline $q_n(x_{1:n})$ is a ``good'' estimator on half of the data,
$\t q^\Slim(x_{1:n})=\text{Bad}(x_{1:n})$ is ``bad'' on all data.
For example, $\text{Bad}(x_{1:n}):=|\X|^{-n}$ (see {\em Uniform} next Section)
and $\text{Good}(x_{1:n})={n+d-1\choose n_1...n_d~d-1}$ (see {\em Laplace} next Section)
or simpler $\text{Good}(1_{1:n})=1$, lead to $R_n(\t q^\Slim)\propto n$.

\paradot{Mixture}
Another way of exploiting $\bar q_s$ is as follows:
Rather than taking the limit $s\to\infty$ let us consider the class $\{\bar q_1,\bar q_2,...\}$ of {\em all}
$\bar q_s$. This corresponds to a set of measures on $\X^\infty$, each good in a particular
circumstance, namely $\bar q_s$ is good and indeed perfect at time $s$.
It is therefore natural to consider a Bayesian mixture over this class \cite{Santhanam:06}
\beq\label{eq:qmix}
  \t q^\Smix(x_{1:n}) ~:=~ \sum_{s=0}^\infty\bar q_s(x_{1:n}) w_s
  \qmbox{with prior} w_s>0, ~~\sum_{s=0}^\infty w_s=1.
\eeq
$\t q^\Smix$ is \ref{TC} and its regret can easily be upper bounded \cite{Santhanam:06}:
\beq\label{eq:lnwbnd}
  R_n(\t q^\Smix) ~=~ \max_{x_{1:n}}\ln {q_n(x_{1:n})\over\sum_{s=0}^\infty \bar q_s(x_{1:n}) w_s}
  ~\leq \max_{x_{1:n}}\ln {q_n(x_{1:n})\over\bar q_n(x_{1:n}) w_n} ~=~ \ln w_n^{-1}
\eeq
For e.g.\ $w_n:={1\over (n+1)(n+2)}$ we have $\ln w_n^{-1}\leq 2\ln(n+2)$
which usually can be regarded as small.
This shows that {\em any} offline estimator can be converted into
an online predictor with very small extra regret \req{eq:ExR}. Note
that while $\t q^\Smix$ depends on arbitrary $Q$ defined in
\req{eq:qbar}, the upper bound \req{eq:lnwbnd} on $R_n$ does not.
Unfortunately it is unclear how to convert this heavy construction
into an efficient algorithm.

A variation is to set $Q\equiv 0$, which makes $\t q^\Smix$ a
semi-measure, which could be made TC by naive normalization \req{eq:qn1}. Bound
\req{eq:lnwbnd} still holds since for $\t q^\Smix$ with $Q\equiv 0$ the
normalizer $\N\leq 1$.
Another variation is as follows.
Often $q_n$ violates \ref{TC} only weakly, in
which case a sparser prior, e.g. $w_{2^k}:={1\over(k+1)(k+2)}$ and
$w_n=0$ for all other $n$, can lead to even smaller regret.

\paradot{Further choices for $\t q$}
Of course the four presented choices for $\t q$ do not exhaust all
options. Indeed, finding a tractable $\t q$ with good properties is
a major open problem. Several estimation procedures do not only
provide $q_n$ on $\X^n$, but measures on $\X^\infty$ or
equivalently for {\em each} $n$ {\em separately} a \ref{TC}
$q_n:\X^*\to[0;1]$ (see Bayes and crude MDL below). While this
opens further options for $\t q$, e.g.\ $\t
q(x_{n+1}|x_{1:n}):=q_n(x_{1:n+1})/q_n(x_{1:n})$ with some (weak)
results for MDL \cite{Hutter:05mdl2px}, it does not solve our main
problem.

\paradot{Notes}
Each solution attempt has its down-sides, and
a solution satisfying all our criteria remains open.

It is easy to verify that, if $q_n$ is already \ref{TC}, the first
three definitions of $\t q$ coincide, and $R_n=0$, which is
reassuring, but $\t q_n^\Smix$ in general differs due to the
arbitrary $w$ in \req{eq:qmix} and arbitrary $Q$ in $\bar q$ in \req{eq:qbar}.

\section{Examples}\label{sec:Ex}

All examples below fall in one of two major strategies for
designing estimators (the introduction mentions others we do not
consider). One strategy is to start with a class $\M$ of
probability measures $\nu$ on $\X^\infty$ in the hope one of them
is good. For instance, $\M$ may contain (a subset of) i.i.d.\
measures
$\nu_{\v\theta}(x_{1:n}):=\theta_{x_1}\cdot...\cdot\theta_{x_n}$
with $\theta_i\geq 0$ and $\theta_1+...+\theta_d=1$ and $d:=|\X|$.
One may either select a $\nu$ from $\M$ informed by given data
$x_{1:n}$ or take an average over the class. The other strategy
assigns uniform probabilities over subsets of $\X^n$. This
combinatorial approach will be described later. Some strategies
lead to \ref{TC} and some examples are \ref{TC}. For the others we
will discuss the various online conversions $\t q$.

\paradot{Bayes}
The Bayesian mixture over $\M$ w.r.t.\ some
prior (density) $w()$ is defined as
\beqn
  q_n(x_{1:n}) ~:=~ \int_\M \nu(x_{1:n})\,w(\nu)\,d\nu
\eeqn
Since $q_n$ is \ref{TC},
$(q_n^\Srat)\equiv(q_n^\Sno)\equiv(q_n^\Slim)$ coincide with $\t q$,
$R_n=0$, and $\t q^\Srat$ is tractable if the Bayes mixture is.
Note that $\t q\not\in\M$ in general, in particular it is not
i.i.d. Assume the true sampling distribution $\mu$ is in $\M$. For
countable $\M$ and counting measure $d\nu$, we have
$q_n(x_{1:n})\geq\mu(x_{1:n})w(\mu)$, hence
$R_n^{\text{online}}=R_n^{\text{offline}}\leq\ln w(\mu)^{-1}$. For
continuous classes $\M$ we have
$R_n^{\text{online}}=R_n^{\text{offline}}\lesssim\ln w(\mu)^{-1}+O(\ln n)$ under some mild conditions
\cite{Barron:91,Hutter:03optisp,Hutter:07pquest}.

\paradot{MDL/NML/MAP}
The MAP or MDL estimator is
\beqn
  \hat q_n(x_{1:n}) ~:=~ \sup_{\nu\in\M}\{\nu(x_{1:n})\,w(\nu)\}
  \qmbox{and} q_n(x_{1:n}) ~:=~ {\hat q_n(x_{1:n})\over\sum_{x_{1:n}}\hat q_n(x_{1:n})}
\eeqn
Since $\hat q_n$ is not even a probability on $\X^n$, we have to
normalize it to $q_n$. For uniform prior density $w()$, $\hat q_n$
is the maximum likelihood (ML) estimator, and $q_n$ is known under
the name normalized maximum likelihood (NML) or modern minimum description length (MDL).
Unlike Bayes, $q_n$ is {\em not} \ref{TC}, which causes all kinds of
complications
\cite{Gruenwald:07book,Hutter:09mdltvp,Hutter:14martoscx},
many of them can be traced back to our main open problem
and the unsatisfactory choices for $\t q$ \cite{Hutter:05mdl2px}.
$R_n^{\text{offline}}$ is essentially the same as for Bayes under similar conditions,
but $R_n^{\text{online}}$ depends on the choice of $\t q$.
Crude MDL simply selects
$q_n:=\arg\max_{\nu\in\M}\{\nu(x_{1:n})\,w(\nu)\}$ at time $n$,
which is a probability measure on $\X^\infty$. While this opens
additional options for defining $\t q$, they also can perform
poorly in the worst case \cite{Hutter:05mdl2px}.
Note that most versions of MDL perform often very well in practice,
comparable to Bayes; robustness and proving guarantees are the open problems.

\paradot{Uniform}
The uniform probability $q_n(x_{1:n}):=|\X|^{-n}$ is \ref{TC}, hence all
four $\t q$ coincide and $R_n=0$ (only for uniform $Q$ in case of $q_n^\Smix$).
Unless data is uniform, this is a
lousy estimator, since predictor $\t q(x_t|x_{<t})=1/|\X|$ is
indifferent and ignores all evidence $x_{<t}$ to the contrary.
But the basic idea of uniform probabilities is sound, if applied
smartly: The general idea is to partition the sample space (here $\X^n$)
into $\P=\{S_1,...,S_{|\P|}\}$ and assign uniform probabilities to each partition:
$q_n(x_{1:n}|S_r)=1/|S_r|$ and a (possibly) uniform probability to the parts themselves
$q_n(S_r)=1/|\P|$. For small $|\P|$, $q_n(x_{1:n})=q_n(x_{1:n}|S_r)q_n(S_r)$ is never more than a small factor
$|\P|$ smaller than uniform $|\X|^{-n}$ but may be a huge factor of $|\X|^n/|S_r||\P|$ larger.
The Laplace rule can be derived that way, and the Good-Turing and Ristad estimators
by further sub-partitioning.

\paradot{Laplace}
More interesting than the uniform probability is the following
double uniform combinatorial probability: Let $n_i:=|\{t:x_t=i\}|$
be the number of times, symbol $i\in\X=\{1,...,d\}$ appears in
$x_{1:n}$. We assign a uniform probability to all sequences
$x_{1:n}$ with the same counts $\v n:=(n_1,...,n_d)$, therefore
$q_n(x_{1:n}|\v n)={n\choose n_1...n_d}^{-1}$. We also assign a
uniform probability to the counts $\v n$ themselves, therefore
$q_n(\v n)=|\{\v n:n_1+...+n_d=n\}|^{-1}={n+d-1\choose d-1}^{-1}$.
Together
\beqn
  q_n(x_{1:n}) ~=~ {n\choose n_1~...~n_d}^{-1} {n+d-1\choose d-1}^{-1} ~=~ {n+d-1\choose n_1~...~n_d~~d-1}^{-1}
\eeqn
\beqn
  \text{hence}~~\t q^\Srat(x_{n+1}=i|x_{1:n}) ~=~ {q_{n+1}(x_{1:n}i)\over q_n(x_{1:n})} ~=~ {n_i+1\over n+d}
\eeqn
is properly normalized (\ref{Norm}), so $\t q^\Srat$ is \ref{TC}, and
$(q_n^\Srat)\equiv(q_n^\Sno)\equiv(q_n^\Slim)$ coincide with $\t q$ and $R_n=0$.
$\t q^\Srat$ is nothing but Laplace's famous rule.

\paradot{Good-Turing}
Even more interesting is the following triple uniform probability:
Let $M_r:=\{i:n_i=r\}$ be the symbols that appear exactly
$r\in\SetN_0$ times in $x_{1:n}$, and $m_r:=|M_r|$ be their number.
Clearly $m_r=0$ for all $r>n$, but due to $\sum_{r=0}^n r\cdot
m_r=n$, $m_r=0$ also for many $r<n$. We assign uniform
probabilities to $q_n(x_{1:n}|\v n)$ as before and to $q_n(\v n|\v
m)$ and to $q_n(\v m)$, where $\v m:=(m_0,...,m_n)$. There are
${d\choose m_0...m_n}$ ways to distribute symbols
$1,...,d$ into sets $(M_0,...,M_n)$ (many of them empty) of sizes
$m_0,...,m_n$. Therefore $q_n(\v n|\v m)={d\choose
m_0...m_n}^{-1}$. Each $\v m$ constitutes a decomposition of $n$
into natural summands with repetition but without regard to order.
The number of such decompositions is a well-known function \cite[\S 24.2.2]{Abramowitz:74}
which we denote by $\text{Part}(n)$. Therefore $q_n(\v
m)=\text{Part}(n)^{-1}$. Together
\bqa
  \label{eq:uuu} q_n(x_{1:n}) &=& {n\choose n_1~...~n_d}^{-1} {d\choose m_0~...~m_n}^{-1} \text{Part}(n)^{-1},~~~\text{hence} \\
  \label{eq:uuuc} \t q^\Srat(x_{n+1}=i|x_{1:n}) &=& {q_{n+1}(x_{1:n}i)\over q_n(x_{1:n})} = {n_i+1\over n+1}\cdot{m_{r+1}+1\over m_r}\cdot{\text{Part}(n)\over\text{Part}(n+1)},~r=n_i~~~~~~~
\eqa
This is not \ref{TC} as can be verified by example, but is a very interesting predictor:
The first term is close to a frequency estimate $n_i/n$.
The second term is close to the Good-Turing (GT) correction $m_{r+1}/m_r$.
The intuition is that if e.g.\ many symbols have appeared once
($m_1$ large), but few twice ($m_2$ small), we should be skeptical
of observing a symbol that has been observed only once another time,
since it would move from a likely category to an unlikely
one. The third term ${\text{Part}(n)\over\text{Part}(n+1)}\to 1$ for $n\to\infty$.
The normalized version
\bqa \label{eq:GTn1}
  \nq\t q^\Sno(x_{n+1}=i|x_{1:n}) &=& {\t q^\Srat(x_{n+1}=i|x_{1:n}) \over \sum_{x_{n+1}} \t q^\Srat(x_{n+1}|x_{1:n})}
  ~=~ {1\over\N_n}\cdot{r+1\over n+1}\cdot{m_{r+1}+1\over m_r}
\\ \label{def:GTNn}
  \qmbox{where} \N_n &:=& {1\over n+1}\sum_{r=0,m_r\neq 0\nq\nq}^n(r+1)(m_{r+1}+1)
\eqa
is even closer to the GT estimator. We kept ${1\over n+1}$ as in \cite[Eq.(13)]{Good:53},
while often ${1\over n}$ is seen due to \cite[Eq.(2)]{Good:53}.
Anyway after normalization there is no difference. The only
difference to the GT estimator is the appearance of $m_{r+1}+1$
instead of $m_{r+1}$. Unfortunately its regret is very large:

\begin{theorem}[Naively normalized triple uniform estimator]\label{thm:Rn1uuu}
Naive normalization of the triple uniform combinatorial offline
estimator $q_n$ defined in \req{eq:uuu} leads to the (non-smoothed)
Good-Turing estimator $\t q^\Sno$ given in \req{eq:GTn1} with regret
\beq\label{eq:RnGT}
  R_n(\t q^\Sno||q_n) = \max_{x_{1:n}}\Big\{\sum_{t=1}^n\ln\N_{t-1}\Big\} - \ln(\text{Part}(n))~
  \left\{ { = n\ln 2 \pm O(\sqrt{n}) ~\text{for}~ |\X|=\infty
            \atop \geq 0.43n-O(\sqrt{n}) ~\text{for}~ |\X|\geq 3 } \right.
\eeq
\end{theorem}

Inserting \req{eq:uuuc} and \req{def:GTNn} into \req{eq:qn1c} we
get $\N(x_{1:n})={\t q^\Srat(x_{n+1}|x_{1:n})\over\t
q^\Sno(x_{n+1}|x_{1:n})} =
{\text{Part}(n)\over\text{Part}(n+1)}\N_n$ which by \req{eq:Rnqno}
implies the first equality.
We prove the last equality in Appendix~\ref{app:RnGT} by showing
that the maximizing sequence is $x_{1:\infty}=1223334444...$
with $\N_n=2\pm O(n^{-1/2})$ which requires infinite $d$ or at
least $d\geq\sqrt{2n}$.
We also show that $R_n\geq 0.43n-O(\sqrt{n})$ for every
$d\geq 3$. The linearly growing $R_n$ shows that naive
normalization severely harms the offline triple uniform estimator
$q_n$.

Indeed, raw GT performs very poorly for large $r$ in practice, but
smoothing the function $m_{()}$ leads to an excellent estimator in
practice \cite{Good:53}, e.g.\ Kneser-Ney smoothing for text data
\cite{Chen:99}. Our $m_{r+1}\leadsto m_{r+1}+1$ is a kind of albeit
insufficient smoothing. $\t q^\Smix$ may be regarded as an
(unusual) kind of smoothing, which comes with the strong guarantee
$R_n\leq 2\ln(n+2)$, but a direct computation is prohibitive.
\cite{Santhanam:06} gives a low-complexity smoothing of the
original GT that comes with guarantees, namely sub-linear
$O(n^{2/3})$ log worst-case sequence attenuation, but this is
different from $R_n$ in various respects: Log worst-case
sequence-attenuation is relative to i.i.d.\ coding and unlike $R_n$
lower bounded by $O(n^{1/3})$. Still a similar construction may
lead to sublinear and ideally logarithmic $R_n$.

\paranodot{Ristad}
\cite{Ristad:95} designed an interesting quadruple uniform probability motivated as follows:
If $\X$ is the set of English words and $x_{1:n}$ some typical
English text, then most symbols=words will not appear ($d\gg n$). In this
case, Laplace assigns not enough probability (${n_i+1\over
n+d}\ll{n_i\over n}$) to observed words. This can be rectified by
treating symbols $\A:=\{i:n_i>0\}$ that do appear different from
symbols $\X\setminus\A$ that don't.
For $n>0$, $x_{1:n}$ may contain $m\in\{1,...,\min\{n,d\}\}$ different symbols, so we set
$q_n(m)=1/\min\{n,d\}$.
Now choose uniformly which $m$ symbols $\A$ appear, $q_n(\A|m)={d\choose m}^{-1}$ for $|\A|=m$.
There are ${n-1\choose m-1}$ ways of choosing the frequency of symbols consistent with
$n_1+...+n_d=n$ and $n_i>0\Leftrightarrow i\in\A$, hence $q_n(\v n|\A)={n-1\choose m-1}^{-1}$.
Finally, $q_n(x_{1:n}|\v n)={n\choose n_1...n_d}^{-1}$ as before.
Together
\beq\label{def:uuuu}
  q_n(x_{1:n}) ~=~ {n\choose n_1~...~n_d}^{-1} {n-1\choose m-1}^{-1} {d\choose m}^{-1} {1\over\min\{n,d\}},
  \qmbox{which implies}
\eeq
\beqn
  \t q^\Srat(x_{n+1}=i|x_{1:n}) ~=~ {\min\{n,d\}\over\min\{n+1,d\}}\cdot
  \left\{
  { {(n_i+1)(n-m+1)\over n(n+1)}          \qmbox{if} n_i>0   \atop
    {m(m+1)\over n(n+1)}\cdot{1\over d-m} \qmbox{if} n_i=0 }
  \right.
\eeqn
This is not \ref{TC}, since
\beqn
  \N(x_{1:n}) ~=~ {\min\{n,d\}\over\min\{n+1,d\}}\cdot
  \left\{
  { 1+{2m\over n(n+1)}~     \qmbox{if} m<d   \atop
    1-{m(m-1)\over n(n+1)} \qmbox{if} m=d }
  \right.
\eeqn
is not identically 1. Normalization leads to
\beq\label{eq:equuuun1}
  \t q^\Sno(x_{n+1}=i|x_{1:n}) ~=~
  \left\{
    \begin{array}{ccl}
      {(n_i+1)(n-m+1)\over n(n+1)+2m} &          ~\text{if}~ & n_i>0 ~\text{and}~ m<d \\
      {m(m+1)\over n(n+1)+2m}\cdot{1\over d-m} & ~\text{if}~ & n_i=0 \\
      {n_i+1\over n+m} &                         ~\text{if}~ & m=d~~[\Rightarrow n_i>0]
    \end{array}
  \right.
\eeq
For $n=0$ we have $\t q^\Srat(x_1)=\t q^\Sno(x_1)=q_n(x_1)=1/d$ and $\N(\epstr)=1$.
While by construction, the offline estimator should have good performance
(in the intended regime), the performance of the online version depends on how much
the normalizer exceeds 1. The first factor in $\N$ is $\leq 1$ and the $m=d$ case is $\leq 1$.
Therefore $\N(x_{1:n})\leq 1+{2m\over n(n+1)}\leq
1+{2\over n+1}$, where we have used $m\leq n$ in the second step.
The regret can hence be bounded by
\beqn
  R_n(\t q^\Sno) ~\leq~ \sum_{t=1}^n \ln\max_{x_{<t}}\N(x_{<t})
  ~\leq~ \sum_{t=2}^n \ln(1\!+\!\fr2t)
  ~\leq~ \sum_{t=2}^n \fr2t
  ~\leq~ 2\ln n
\eeqn

\begin{theorem}[Quadruple uniform estimator]\label{thm:Rn1uuuu}
Naive normalization of Ristad's quadruple uniform combinatorial offline
estimator $q_n$ defined in \req{def:uuuu} leads to
Ristad's natural law $\t q^\Sno$ given in \req{eq:equuuun1} with regret
$R_n(\t q^\Sno||q_n)\leq 2\ln n$.
\end{theorem}

This shows that simple normalization does not ruin performance.
Indeed, the regret bound is as good as we are able to guarantee in
general via $\t q^\Smix$.

\section{Computational Complexity of $\t q$}\label{sec:CCqt}

\paradot{Computability and complexity of $\t q^\Smix$}
From the four discussed online estimators only $q_n^\Smix$
guarantees small extra regret over offline $(q_n)$ in general, but
the definition of $\t q^\Smix$ is quite heavy and at first it is
not even clear whether it is computable. The following theorem
shows that $\t q^\Smix$ can be computed to relative accuracy $\eps$
in double-exponential time:

\begin{theorem}[Computational complexity of $\t q^\Smix$]\label{thm:ccqtmix}
There is an algorithm $A$ that computes $\t q^\Smix$ (with uniform choice for $Q$)
to relative accuracy $|A(x_{1:n},\eps)/\t q^\Smix(x_{1:n})-1|<\eps$ in time
$O(|\X|^{4|\X|^n\!/\eps})$ for all $\eps>0$.
\end{theorem}

The relative accuracy $\eps$ allows us to
compute the predictive distribution $\t q^\Smix(x_t|x_{<t})$ to
accuracy $\eps$, ensures $A(x_{1:n},\eps)>(1-\eps)\t
q_n^\Smix(x_{1:n})$, hence $R_n(A(\cdot,\eps)||q_n)\leq R_n(\t
q^\Smix||q_n)+{\eps\over 1-\eps}$, and approximate
normalization $|1-\sum_{x_{1:n}}A(x_{1:n},\eps)|<\eps$.

\paradot{Computational complexity of general $\t q$}
The existence of $\t q^\Smix$ shows that any offline estimator can
be converted into an online estimator with minimal extra regret
$R_n\leq 2\ln(n+2)$. While encouraging and of theoretical interest,
the provided algorithm for $\t q^\Smix$ is prohibitive. Indeed,
Theorem~\ref{thm:subopt} below establishes that there exist offline
$(q_n)$ computable in polynomial time for which the fastest
algorithm for {\em any} online (=TC) $\t q$ with $R_n\leq O(\log
n)$ is at least exponential in time.

Trivially $R_n\leq n\ln|\X|$ can always be achieved for any $(q_n)$
by uniform $\t q(x_{1:n})=|\X|^{-n}$. So a very modest quest would
be $R_n\leq (1-\eps)n\ln|\X|$. If we require $\t q$ to run in
polynomial time but with free oracle access to $(q_n)$,
Theorem~\ref{thm:poor} below shows that this is also not possible
for some exponential time $(q_n)$.

Together this does not rule out that for every fast $(q_n)$ there
exists a fast $\t q$ with e.g. $R_n\leq\sqrt{n}$. This is our main
remaining open problem to be discussed in Section~\ref{sec:Open}.

The main proof idea for both results is as follows: We construct a
deterministic $(q_s)$ that is 1 on the sequence of quasi-independent quasi-random
strings $\dot x_{1:1}^1$, $\dot x_{1:2}^2$, $\dot x_{1:3}^3$,~...~.
The only way for $\t q(x_{1:n})$ to be not too much smaller than $\bar
q_s(\dot x_{1:n}^s)$ is to know $\dot x_{1:s}^s$. If $s=s(n)$ is
exponential in $n$ this costs exponential time. If $\t q$ has only
oracle access to $(q_s)$, it needs exponentially many oracle calls
even for linear $s(n)=(1+\eps)n$.

The general theorem is a bit unwieldy and is stated and proven in
the next section. Here we present and discuss the most interesting
special cases.
%
$\DTIME(g(n))$ is defined as the class of all algorithms that run
in time $O(g(n))$ on inputs of length $n$. Real-valued algorithms
produce for any rational $\eps>0$ given as an extra argument, an
$\eps$-approximation in this time, as did $A(x_{1:n},\eps)$ for $\t
q^\Smix$ above. Algorithms in $\text{E}^c:=\DTIME(2^{cn})$ run in
exponential time, while $\text{P}:=\bigcup_{k=1}^\infty\DTIME(n^k)$
is the classical class of all algorithms that run in polynomial
time (strictly speaking Function-P or FP \cite{Arora:09}).
%
The theorems don't rest on any complexity separation
assumptions such as P$\neq$NP.
We only state and prove the theorems for binary alphabet
$\X=\SetB=\{0,1\}$. The generalization to arbitrary finite alphabet
is trivial. `For all large $n$' shall mean `for all but finitely
many $n$', denoted by $\forall'n$. $m>0$ is a constant that depends
on the machine model, e.g. $m=1$ for a random access machine (RAM).

\begin{theorem}[Sub-optimal fast online for fast offline]\label{thm:subopt}
For all $r>0$ and $c>0$ and $\eps>0$
\bqan
  & (i) & \exists(q_s)\in\DTIME(s^{b+m})~
  \forall\t q\in\text{E}^c: R_n(\t q||q_n)\geq r\ln n~\forall'n,
  ~\text{where}~ b:=\fr{c+1+\eps}{1-\eps}r
\\
  & (ii) & \text{in particular for large $c$ and $r$:}~ \exists(q_s)\in\text{P}~
  \forall\t q\in\text{E}^c: R_n\geq r\ln n~\forall'n
\\
  & (iii) & \text{in particular for small $c,\eps$:}~ \exists(q_s)\!\in\!\DTIME(s^{r+m+\eps})\,
  \forall\t q\!\in\!\text{P}: R_n\geq r\ln n~\forall'n
\\
  & (iv) & \text{in particular for $\t q^\Smix$:}~~~ \exists(q_s)\in\text{P} : \t q^\Smix\not\in\text{E}^c
\eqan
\end{theorem}
In particular (iii) implies that there is an offline estimator
$(q_s)$ computable in quartic time $s^4$ on a RAM for which no
polynomial-time online estimator $\t q$ is as good as $\t q^\Smix$.
The slower $(q_s)$ we admit (larger $r$), the higher the lower
bound gets. (ii) says that even algorithms for $\t q$ running in
exponential time $2^{cn}$ cannot achieve logarithmic regret for all
$(q_s)\in\text{P}$. In particular this implies that (iv) any
algorithm for $\t q^\Smix$ requires super-exponential time for some
$(q_s)\in\text{P}$ on some arguments.

The next theorem is much stronger in the sense that it rules out
even very modest demands on $R_n$ but is also much weaker since it
only applies to online estimators for slow $(q_s)$ used as a black box
oracle. That is, $\t q^o(x_{1:n})$ can call $q_s(z_{1:s})$ for any
$s$ and $z_{1:s}$ and receives the correct answer. We define
$\DTIME^o(g(n))$ as the class of all algorithms with such oracle
access that run in time $O(g(n))$, where each oracle call is
counted only as one step, and similarly $\text{P}^o$ and $\text{E}^{c,o}$.

\begin{theorem}[Very poor fast online using offline oracle]\label{thm:poor}
For all $\eps>0$
\bqan
  & & \exists o\equiv(q_s)\in\text{E}^1~
  \forall \t q^o\in\text{E}^{\eps/2,o}:
  R_n(\t q^o||q_n)\geq (1-\eps)n\ln 2~\forall'n
\\
  & & \text{or cruder:}~\exists o\equiv(q_s)~
  \forall \t q^o\in\text{P}^o:
  R_n(\t q^o||q_n)\geq (1-\eps)n\ln 2~\forall'n
\eqan
\end{theorem}
The second line states that the trivial bound $R_n\leq n\ln 2$ achieved
by the uniform distribution can in general not be improved by a fast
$\t q^o$ that (only) has oracle access to the offline estimator.

Usually one Does not state the complexity of the oracle, since it
does not matter, but knowing that an $o\in\text{E}^1$ is sufficient
(first line) tells us something: First, the negative result is not
an artifact of some exotic non-computable offline estimator. On the
other hand, if an exponential time offline $o$ is indeed needed to
make the result true, the result wouldn't be particularly
devastating. It is an open question whether an $o\in\text{P}$ can
cause such bad regret.

\section{Computational Complexity Proofs}\label{sec:CCqtproofs}

\paradot{Proof of Theorem~\ref{thm:ccqtmix}}
The design of an algorithm for $\t q^\Smix$ and the analysis of its
run-time follows standard recipes, so will only be sketched. A
real-valued function $\t q^\Smix:\X^*\to[0;1]$ is (by definition)
computable (also called estimable
\cite{Hutter:04uaibook}), if there is an always halting
algorithm $A:\X^*\times\SetQ^+\to\SetQ$ with $|A(x_{1:n},\eps)-\t
q^\Smix(x_{1:n})|<\eps$ for all rational $\eps>0$. We assume there
is an oracle $q_t^{\eps}$ that provides $q_t$ to $\eps$-accuracy in
time $O(1)$. We assume that real numbers can be processed in unit
time. In reality we need $O(\ln\frs{1}\eps)$ bits to represent, and
time to process, real numbers to accuracy $\eps$. This leads to
some logarithmic factors in
run-time which are dwarfed by our exponentials, so will be ignored.
To compute $\bar q_s(x_{1:n})$ to accuracy $\frs{\eps}{2}$ we need
to call $q_s^{\eps/2N}$ oracle $N:=\max\{|\X|^{s-n},1\}$ times and
add up all numbers. We can compute $\t q^\Smix$ to $\eps$-accuracy
by the truncated sum $\sum_{s=0}^{2/\eps}\bar
q_s^{\eps/2}(x_{1:n})w_s$ with $w_s={1\over(s+1)(s+2)}$, since the
tail sum is bounded by $\frs{\eps}{2}$. Hence overall runtime is
$O(|\X|^{2/\eps-n})$. But this is not sufficient. For large $n$,
$\t q^\Smix(x_{1:n})$ is typically small, and we need a {\em relative}
accuracy of $\eps$, i.e.\ $|A(x_{1:n},\eps')/\t
q^\Smix(x_{1:n})-1|<\eps$.
For $Q(x_{1:n})=|\X|^{-n}$, we have $\t q^\Smix(x_{1:n})\geq
\fr12 Q(x_{1:n})=\fr12|\X|^{-n}$, hence $\eps'=\fr{\eps}{2}|\X|^{-n}$
suffices. Run time becomes $O(|\X|^{{4\over\eps}|\X|^n-n})\leq
\e^{\e^{O(n)}/\eps}$.
\qed

\begin{theorem}[Fast offline can imply slow online (general)]\label{thm:foso}
Let $s(n)$ and $f(n)$ and $g(n)$ be monotone increasing functions.
$s(n)$ shall be injective and $\geq n$ for large $n$ with inverse
$n(s):=\max\{n:s(n)\leq s\}$ and $g(n)<\fr12 n^{-\d}h(n)$, where
$h(n):=2^{s(n)-n}[n^{-\g}-2^{f(s(n))-n}]$.
$m>0$ is a constant depending on the machine model, e.g. $m=1$ for a RAM.
Then for all $\g>0$ and
$\delta>0$ it holds that
\bqan
  & & \exists o\equiv(q_s)\in\DTIME(n(s)^{\g+\delta}2^{n(s)}g(n(s))s^m)~
\\
  & & \forall \t q^o\in\DTIME^o(g(n)):
  R_n(\t q^o||q_n)\geq f(n)\ln 2~\forall'n
\eqan
\end{theorem}

\paradot{Proof of Theorem~\ref{thm:foso}}

\paradot{Effective quasi-sparse sets}
We need a single set $\{\dot x_{1:s}^s\}_{s\in\SetN}\equiv\{\dot
x_1^1,\dot x_{1:2}^2,\dot x_{1:3}^3,...\}$ of sequences that is
``safe'' against {\em every} polynomial time $\t q$ in a sense to
be clarified below. Let $o=(q_s)$ be any deterministic oracle, i.e.
for every $s$, $q_s$ is 1 on exactly one string, namely $\dot
x_{1:s}^s$. Let $T_1^o,T_2^o,...$ be an enumeration of all Turing
machines with access to oracle $o$, but each $T_k^o(z_{1:n})$ is terminated
after time $k^{\delta/\g}g(n)$. Any $\delta>0$ and $\g>0$
will do. Therefore $(T_k^o)$ enumerates all time-bounded machines.

The idea of the following construction is to return an $\dot x_{1:s}^s$ that
is not in any effective quasi-sparse set of the form
\beqn
  L_k^{n,o}:=\left\{
  {\tilde L_k^{n,o} \qmbox{if} |\tilde L_k^{n,o}|\leq 2^{f(s(n))}
        \atop \{\}  ~~\qmbox{else}~~~~~~~~~~~~~~~~~~~~~~ } \right.,
  \qmbox{where} \tilde L_k^{n,o} :=\{z\in\SetB^n:T_k^o(z)=1\}
\eeqn
and $f(s)$ is some (linear/logarithmic) monotone increasing function
and $s(n)$ is some injective (linear/exponential) monotone increasing function.

\paradot{Constructing quasi-random sequences $\dot x_{1:s}^s$}
For the construction to work, $\dot x_{1:s}^s$ should also not be
probed by any fast algorithm on any input. Since the algorithms can
probe oracle $o=(q_s)$ before $q_s$ has been constructed, we need a
careful construction in stages $s=1,2,3,...$. Assume $\dot
x_{1:s'}^{s'}$ and $q_{s'}$ have already been constructed for all
$s'<s$.
We now construct $\dot x_{1:s}^s$. For this we define
a fake oracle $o_s$ that coincides with $o$ whenever queried with a
string of length less than $s$ (the already constructed $q_{s'}$),
but always returns 0 when queried with a string of length $s$ or
larger (for which $q_{s'}$ has yet to be constructed). Let $n:=n(s)$ and
\beqn
  C_{\geq s} ~:=~ \{y_{1:s'} : s'\geq s, \exists z_{1:n}\exists k\leq n^\g : T_k^{o_s}(z_{1:n}) ~\text{calls}~ o_s ~\text{on}~ y_{1:s'} \}
\eeqn
be the set of sequences $y_{1:s'}$ longer or equal than $s$ (this
is important) that are queried by any of the first $n^\g$ (any
$\g>0$ will do) Turing machines $T_k^{o_s}$ on any input
$z_{1:n}$. Now let\vspace{-1ex}
\beqn
  F_s ~:=~ \SetB^s \setminus \Big( \bigcup_{k=1}^{n^\g} L_k^{n,o_s}\times\SetB^{s-n}\cup\bigcup_{s'=1}^s C_{\geq s'} \Big)
\eeqn
be the set of strings of length $s$ that roughly $(i)$ are not
queried and $(ii)$ whose length $n$ prefix is not in any quasi-sparse
set. If $F_s\neq\{\}$,
\beqn
  \text{let} ~\dot x_{1:s}^s ~\text{be the lexicographically first string in}~ F_s
\eeqn
If $F_s=\{\}$, arbitrarily let $\dot x_{1:s}^s=0_{1:s}$.
In any case define $q_s(\dot x_{1:s}^s) := 1$, and 0 on all other sequences of length $s$.

\paradot{Fast good $\t q^o$ implies $F_s=\{\}$}
Let $\t q^o$ be an online (=TC) estimator with access to oracle
$o=(q_s)$ and small regret
\beq\label{eq:Rfbnd}
  R_n(\t q^o||q_n)~\equiv~\max_{x_{1:n}}\ln{q_n(x_{1:n})\over\t q^o(x_{1:n})} ~<~ f(n)\ln 2
  \qmbox{for all large $n$}
\eeq
This implies $\t q^o(x_{1:s}) > 2^{-f(s)}q_s(x_{1:s}) ~\forall
x_{1:s}\forall's$ and in particular $\t q^o(\dot x_{1:s}^s) >
2^{-f(s)}$. Since $\t q^o$ is TC, we have
$\t q^o(\dot x_{1:n}^s) \geq \t q^o(\dot x_{1:s}^s) > 2^{-f(s(n))}$.

Now let us assume that $\t q^o\in\DTIME^o(g(n))$. Then membership in
\beqn
  L^{n,o} ~:=~ \{x_{1:n} : \t q^o(x_{1:n}) > 2^{-f(s)} \}
\eeqn
can be determined in the same (or less) time and
since $\t q^o$ is a probability, $|L^{n,o}|<2^{f(s)}$.
Therefore, there is a $k_0$ such that $T_{k_0}^o$ computes $L^{n,o}=L_{k_0}^{n,o}$.

Now assume $F_s\neq\{\}$ for some $n^\g\geq k_0$.
The construction of $\dot x_{1:s}^s$ is such that
$T_k^{o_s}(z_{1:n})=T_k^o(z_{1:n})$ for all $k\leq n^\g$ and all
$z_{1:n}$, since their oracles coincide
on all queried strings $y_{1:s'}$:
For $y_{1:s'}\neq\dot x_{1:s'}^{s'}$ both oracles answer 0.
For $s'<s$ both oracles coincide also on $\dot x_{1:s'}^{s'}$, and for $s'\geq s$ any queried
string is added to the tabu list $C_{\geq s}$ and the choice of
$\dot x_{1:s'}^{s'}$ outside $\bigcup_{s'=1}^s C_{\geq s'}$ ensures it
has also not been queried earlier in the construction.
So $o$ also returns 0 for $s'\geq s$ on all queried strings.

In particular $T_{k_0}^{o_s}(z_{1:n})=T_{k_0}^o(z_{1:n})$, hence
$L_{k_0}^{n,o}=L_{k_0}^{n,o_s}$. Further, $\dot x_{1:s}^s\in F_s$
implies prefix $\dot x_{1:n}^s\not\in L_{k_0}^{n,o_s}$ by definition of $F_s$.
We conclude $\dot x_{1:n}^s\not\in L^{n,o}$, which clearly
contradicts $\t q^o(\dot x_{1:n}^s) > 2^{-f(s)}$. Therefore, $\t
q^o\in\DTIME^o(g(n))$ implies $F_s=\{\}$ for all large $s$.

\paradot{$F_s=\{\}$ implies slow good $\t q^o$}
$t(n):=n^\d g(n)$ upper bounds the running time of $T_k^o$ for $k\leq n^\g$.
It also bounds the number of oracle calls in $T_k^o$,
since each oracle call costs at least one step.
Note that $C_{\geq s'}\subseteq C_{\geq s''}$ if $s'\geq s''$ and $n(s')=n(s'')$,
which implies $\bigcup_{s'=1}^s C_{\geq s'} = \bigcup_{n'=1}^n C_{\geq s(n')}$ due to $s(n(s))\leq s$.
Using
\bqan
  & & |L_k^{n,o_s}|\leq 2^{f(s)} \qmbox{and} |C_{\geq s}|\leq n^\g\,2^n t(n) ~~\Rightarrow~~ \Big|\bigcup_{n'=1}^nC_{\geq s(n')}\Big|\leq n^\g\,2^{n+1}t(n)
\\
  & & \qmbox{implies} |F_s| ~\geq~ 2^s - n^\g\,2^{f(s)}2^{s-n} - n^\g 2^{n+1}t(n)
\\
  & & \qmbox{hence} F_s=\{\} \qmbox{implies} 2t(n)\geq 2^{s-n}[n^{-\g}-2^{f(s)-n}] ~=:~ h(n)
\eqan
This contradicts the assumption on $g(n)$ in the theorem, hence
$F_s\neq\{\}$, hence $\t q^o\not\in\DTIME^o(g(n))$ for all $\t
q^o$ with regret \req{eq:Rfbnd}, whose contrapositive is
\beqn
  \forall \t q^o\in\DTIME^o(g(n)): R_n(\t q^o||q_n)\geq f(n)\ln 2~\forall'n
\eeqn

\paradot{Complexity of $(q_s)$}
The construction of $q_s$ requires running $T_k^{o_s}(z_{1:n})$ for
all $z_{1:n}$ for all $k\leq n^\g$, each requiring $k^{\delta/\g}
g(n)\leq t(n)$ steps. Hence
$q_s\in\DTIME^{o_s}(n^\g 2^n t(n))$ where $n=n(s)$. We can
get rid of the self-reference to oracle $o_s$ by considering the
complexity of the iterative construction of $q_{s'}$ and $\dot
x_{1:s'}^{s'}$ for all $s'\leq s$.

Assume we have constructed and stored $\dot x_{1:s'}^{s'}$ for all
$s'<s$. We construct $\dot x_{1:s}^s$ as follows: First note that
oracle $o_s$ in $T_k^{o_s}$ can be eliminated. If queried for
$y_{1:s'}$ for $s'<s$ we simply return 1 iff $y_{1:s'}=\dot
x_{1:s'}^{s'}$, which can be done in time $s'\leq s$, since $\dot
x_{1:s'}^{s'}$ has been pre-computed and stored. If $o_s$ is
queried for $y_{1:s'}$ for $s'\geq s$ the answer was defined to be
0, clearly computable in time $s$.

To efficiently compute $\dot x_{1:s}^s$, we first construct
$U_n:=\bigcup_{k=1}^{n^\g}L_k^{n,o_s}$ in time $O(n^\g 2^n
t(n)s)$. We now make a list
of the lexicographically first $\min\{2^s,n^\g 2^{n+1} t(n)+1\}$
strings of length $s$ whose length $n$ prefix is {\em not} in
$U_n$. Next we cross out all strings queried in the definition of
$C_{\geq s(n')}$ for all $1\leq n'\leq n$ in time $O(n^\g
2^{n+1} t(n)s)$. The lexicographically first string left over can
be found in time $O(n^\g 2^{n+1} t(n)s)$ and will be $\dot
x_{1:s}^s$. Since $|\bigcup_{n'=1}^n C_{\geq s(n')}|\leq n^\g
2^{n+1} t(n)$ at least one string survived elimination, except
$F_s=\{\}$, in which case $\dot x_{1:s}^s=0_{1:s}$. This
shows that $q_s\in\DTIME(n^\g 2^n t(n)s)$ where $n=n(s)$.
This construction assumed a random access machine (RAM). For other
machines, some extra powers of $s$ may be needed with marginal
effect on the results. So in general
\beqn
  (q_s)\in\DTIME(n(s)^{\g+\delta}2^{n(s)}g(n(s))s^m)
\vspace{-3ex}\eeqn
\qed

\paradot{Proof of Theorem~\ref{thm:subopt}}
In Theorem~\ref{thm:foso}, weaken $\DTIME^o\leadsto\DTIME$ and $\t
q^o\leadsto\t q$, and let $s=2^{(1-\eps)n/r}$ and $f(s)=r\log
s=(1-\eps)n$. Then $h(n)=2^{2^{(1-\eps)n/r}-n}[n^{-\g}-2^{-\eps
n}]$, so clearly $g(n):=2^{cn}<\fr12 n^{-\d}h(n)$ for large $n$. For any $\eps>0$ and
sufficiently large $n$ we have
\bqan
  n^{\g+\delta}2^n g(n)s^m &=& s^m 2^{(c+1)n+(\g+\delta)\log n} \\
  &\leq& s^m 2^{(c+1+\eps)n}
  ~=~ s^m 2^{(c+1+\eps){r\log s\over 1-\eps}}
  ~=~ s^{m+{c+1+\eps\over 1-\eps}r} ~=~ s^{b+m}
\eqan
This proves (i). (ii) is just a weaker version of (i) since
$\DTIME(s^{b+m})\subset\text{P}$. (iii) follows from the fact that $b:=r+\eps'$
implies $c>0$ for sufficiently small $\eps>0$, and
$\text{E}^c\supset\text{P}$. (iv) follows from (i) and the fact that
$R_n(\t q^\Smix)\leq 2\ln(n+2)<r\ln n\,\forall'n$ for any $r>2$.
\qed

\paradot{Proof of Theorem~\ref{thm:poor}}
In Theorem~\ref{thm:foso} let $s=(1+\eps)n$ and $f(s)=(1-\eps)s$.
Then $h(n)=2^{\eps n}[n^{-\g}-2^{-\eps^2 n}]$,
so clearly $g(n):=2^{\eps n/2}<\fr12 n^{-\d}h(n)$ for large $n$.
For any $\eps>0$ and sufficiently large $n$ we have
$n^{\g+\delta}2^n g(n)s^m = (1+\eps)^m 2^{n+\eps n/2+(\g+\delta+m)\log n} \leq 2^{(1+\eps)n} = 2^s$.
\qed

\section{Open Problems}\label{sec:Open}

We now discuss and quantify the problems that we raised earlier and are still open.
For some {\em specific} collection $(q_n)$ of probabilities,
does there exist a polynomial-time computable time-consistent $\t q$
with $R_n(\t q||q_n)\leq 2\ln(n+2)\,\forall n$?
Note that $\t q^\Smix$ satisfies the bound, but a direct computation is prohibitive.
So one way to a positive answer could be to find an efficient approximation of $\t q^\Smix$.
If the answer is negative for a specific $(q_n)$ one could try to weaken the requirements on $R_n$.
We have seen that for some, (non-\ref{TC}) $(q_n)$, namely Ristad's, simple normalization $\t q^\Sno$ solves the problem.

A concrete unanswered example are the triple uniform Good-Turing probabilities $(q_n)$.
Preliminary experiments indicate that they and
therefore $\t q^\Smix$ are more robust than current heuristic
smoothing techniques, so a tractable approximation of $\t q^\Smix$ would
be highly desirable. It would be convenient and insightful if such a
$\t q$ had a traditional GT representation but with a smarter
smoothing function $m_{()}$.

The nasty $(q_n)$ constructed in the proof of
Theorem~\ref{thm:foso} is very artificial: It assigns extreme
probabilities (namely 1) to quasi-random sequences. It is unknown
whether there is any offline estimator of practical relevance (such
as Good-Turing) for which no fast online estimator can achieve
logarithmic regret.

An open problem for general $(q_n)$ is as follows:
Does there exist for every $(q_n)$ a polynomial-time algorithm that
computes a time-consistent $\t q$ with $R_n(\t q||q_n)\leq
f(n)\,\forall n$. We have shown that this is not possible for
$f(n)=O(\log n)$ and not even for $f(n)=(1-\eps)n\ln 2$ if $\t q$
has only oracle access to $(q_n)$. This still allows for a positive answer to
the following open problem:

\begin{open}[Fast online from offline with small extra regret]\label{open:fofo}
Can every poly\-nomial-time offline estimator $(q_n)$ be converted to
a polynomial-time online estimator $\t q$ with small regret $R_n(\t q||q_n)\leq
\sqrt{n}\,\forall' n$?
Or weaker: $\forall(q_n)\in\text{P}\,\exists\t q\in\text{P}:R_n=o(n)$? %
Or stronger: $\forall(q_n)\in\text{P}\,\exists\t q\in\text{P}:R_n=O(\log n)^2$?
\end{open}

A positive answer would reduce once and for all the problem of
finding good online estimators to the apparently easier problem of
finding good offline estimators. We could also weaken our notion of
worst-case regret to e.g. expected regret $\E[\ln(q_n/\t q)]$.
Expectation could be taken w.r.t.\ $(q_n)$, but other choices are
possible. Other losses than logarithmic also have practical
interest, but I do not see how this makes the problem easier.

Ignoring computational considerations, of theoretical interest is
whether $O(\log n)$ is the best one can achieve in general, say
$\exists q_n\forall\t q:R_n(\t q)\geq\ln n$, or whether a constant is achievable.

Devising general techniques to upper bound $R_n(\t q^\Sno||q_n)$,
especially if small, is of interest too.

\paradot{Acknowledgements}
Thanks to Jan Leike for feedback on earlier drafts.

\section*{References}\label{sec:Bib}

\addcontentsline{toc}{section}{\refname}
\def\refname{\vspace{-4ex}}
\bibliographystyle{alpha}
\begin{small}

\end{small}

\appendix
\section{Proof of Theorem~\ref{thm:Rn1uuu}}\label{app:RnGT}

For GT we prove $\max_{x_{1:n}}\N_n\to 2$, therefore $\max_{x_{1:n}}\N(x_{1:n})\to 2$ due to
${\text{Part}(n)\over\text{Part}(n+1)}\to 1$ for $n\to\infty$.
We can upper bound \req{def:GTNn} as
\bqan
  (n+1)\N_n &=& \sum_{r=0,m_r\neq 0\nq}^n(r+1)m_{r+1} + \sum_{r=0,m_r\neq 0\nq}^n r ~~ + \sum_{r=0,m_r\neq 0\nq}^n 1
\\
  &\leq& \sum_{r'=1}^{n+1} r'm_{r'} + \sum_{r=0}^n r m_r + |\{r:m_r\neq 0\}|
\\
  &=& n + n + |\{r:m_r\neq 0\}| ~\leq~ 2n+\sqrt{2n}+1
\eqan
$|\{r:m_r\neq 0\}|$ under the constraint $\sum_{r=0}^n r
m_r=n$ is maximized for $m_0=...=m_k=1$ and $m_{k+1}=...=m_n=0$ for
suitable $k$. We may have to set one $m_r=2$ to meet the
constraint. Therefore $n=\sum_{r=0}^n r m_r\geq\sum_{r=0}^k r =
{k(k+1)\over 2}\geq\fr12 k^2$, hence
$|\{r:m_r\neq 0\}|=k+1\leq\sqrt{2n}+1$.

For the lower bound we construct a sequence that attains the upper bound. For instance,
$x_{1:k(k+1)/2}=1223334444~...~k...k$ has $m_1=...=m_k=1$, hence
$x_{1:\infty}=1223334444...$ has $m_1\geq 1,...,m_k\geq 1$ for all $n\geq\fr12 k(k+1)$.
Conversely, for any $n$ we have $m_1\geq 1,...,m_k\geq 1$ with $k:=\lfloor\sqrt{2n}\rfloor-1$.
For the chosen sequence we therefore have
\beqn
  (n+1)\N_n ~\geq~ \sum_{r=0}^{k-1}(r+1)(1+1) ~=~ k(k+1) ~\geq~ 2n-3\sqrt{2n}
\eeqn
The upper and lower bounds together imply $\max_{x_{1:n}}\N_n=2\pm
O(n^{-1/2})$, therefore $\max_{x_{1:n}}\N(x_{1:n})=2\pm
O(n^{-1/2})$ due to
${\text{Part}(n)\over\text{Part}(n+1)}=1-O(n^{-1/2})$ \cite{Abramowitz:74}. Inserting
this into \req{eq:RnGT} gives $R_n=n\ln2\pm O(n^{-1/2})$.

The upper bound holds for any $d$, but the lower bound requires
$d=\infty$ or at least $d\geq\sqrt{2n}$. We now show linear growth of
$R_n$ even for finite $d\geq 3$.
The lower bound is based on the same sequence as used in \cite{Santhanam:06}:
For $x_{1:\infty}=12(132)^\infty$ elementary algebra gives
$\N_n={5\over 3}+{7/3\over n+1}$ and $\N_{n+1}={5\over 3}+{5/3\over
n+2}$ and $\N_{n+2}={4\over 3}+{1\over n+3}$ for $n$ a multiple of
3, hence $\N_n\N_{n+1}\N_{n+2}\geq{100\over 27}$ (except
$\N_0\N_1\N_2=\fr23$). Together with asymptotics
$\ln(\text{Part}(n))\sim\pi\sqrt{2n/3}$
\cite{Abramowitz:74}, this implies that $R_n\geq {n\over
3}\ln{100\over 27} - O(\sqrt{n})$.
\qed

\newpage
\section{List of Notation}\label{app:Notation}

\begin{tabbing}
  \hspace{0.18\textwidth} \= \hspace{0.68\textwidth} \= \kill
  {\bf Symbol }      \> {\bf Explanation}                                                    \\[0.5ex]
  $\equiv$           \> identical, equal by definition, trivially equal                      \\[0.5ex]
  ${n\choose n_1...n_d}$ \> multinomial                                                      \\[0.5ex]
  $n\in\SetN_0$      \> length of sequence                                                   \\[0.5ex]
  $t\in\{1,...,n\}$  \> current ``time''                                                     \\[0.5ex]
  $s\in\SetN$        \> any ``time''                                                         \\[0.5ex]
  $\X=\{1,...,d\}$   \> finite alphabet, $d>1$                                               \\[0.5ex]
  $i,x,x_t\in\X$     \> symbol                                                               \\[0.5ex]
  $x_{t:n}\in\X^{n-t+1}$ \> sequence $x_t...x_n$                                             \\[0.5ex]
  $x_{<t}\in\X^{t-1}$\> sequence of length $t-1$                                             \\[0.5ex]
  $\epstr=x_{1:0}=x_{<1}$ \> empty string                                                    \\[0.5ex]
  $Q$                \> any measure on $\X^\infty$                                           \\[0.5ex]
  $q_n:\X^n\to[0;1]$ \> offline estimated probability mass function                          \\[0.5ex]
  $\bar q_s:\X^*\to[0;1]$ \> extends $q_n$ to any \ref{TC} probability on $\X^*$             \\[0.5ex]
  $\t q:\X^*\to[0;1]$\> online estimator desired to be close to $q_n$                        \\[0.5ex]
  $\t q_{|\X^n}$     \> constrains the domain of $\t q$ to $\X^n$                            \\[0.5ex]
  $\log$, $\ln$      \> binary and natural logarithms, respectively                          \\[0.5ex]
  $\DTIME^o(g(n))$   \> algorithms that run in time $O(g(n))$ with access to oracle $o$      \\[0.5ex]
  $\text{P}:=\bigcup_{k=1}^\infty\DTIME(n^k)$ \> ~~~~~~~~polynomial time algorithms           \\[0.5ex]
  $\text{E}^c:=\DTIME(2^{cn})$ \> ~~~exponential time algorithms (much smaller than EXP or even E!) \\[0.5ex]
  $\SetB:=\{0,1\}$   \> binary alphabet                                                      \\[0.5ex]
  $\forall'n$        \> for all but finitely many $n$, short `for all large $n$'             \\[0.5ex]
  quasi              \> akin to but not necessarily an established definition                \\[0.5ex]
\end{tabbing}

\end{document}